\documentclass[runningheads]{llncs}

\usepackage{graphicx}
\usepackage[utf8]{inputenc}
\usepackage[T5]{fontenc}
\usepackage{tabulary}
\usepackage{latexsym}
\usepackage{float}
\usepackage{makecell}
\usepackage{subcaption,booktabs}
\usepackage{microtype}
\usepackage{multirow}
\usepackage{threeparttable} 
\usepackage{amsmath}
\usepackage{tablefootnote}
\usepackage{pifont}
\usepackage{float}
\usepackage{caption}
\usepackage{subcaption}
\usepackage{placeins}
\usepackage{array}
\usepackage{enumerate}
\usepackage[bookmarks,bookmarksopen,bookmarksdepth=3, bookmarksnumbered]{hyperref}
 
\usepackage{amssymb}
\usepackage{pifont}

\usepackage{pgfplots}
\pgfplotsset{width=10cm,compat=1.9}
\begin{document}

\title{XLMRQA: Open-Domain Question Answering on Vietnamese Wikipedia-based Textual Knowledge Source}

\author{Kiet Van Nguyen\inst{1,2,3} \and Phong Nguyen-Thuan Do\inst{1,2,4} \and Nhat Duy Nguyen\inst{1,2,4} \and Tin Van Huynh\inst{1,2,3} \and Anh Gia-Tuan Nguyen\inst{1,2,3} \and Ngan Luu-Thuy Nguyen\inst{1,2,3}}

\institute{University of Information Technology, Ho Chi Minh, Vietnam \and
Vietnam National University, Ho Chi Minh City, Vietnam \and
\email{\{kietnv,tinhv,ngannlt,anhngt\}@uit.edu.vn} \and \email{\{18520126,18520118\}@gm.uit.edu.vn}}

\authorrunning{Nguyen et al.}
\titlerunning{XLMRQA: Open-Domain Question Answering for the Vietnamese language}
\maketitle

\begin{abstract}
Question answering (QA) is a natural language understanding task within the fields of information retrieval and information extraction that has attracted much attention from the computational linguistics and artificial intelligence research community in recent years because of the strong development of machine reading comprehension-based models. A reader-based QA system is a high-level search engine that can find correct answers to queries or questions in open-domain or domain-specific texts using machine reading comprehension (MRC) techniques. The majority of advancements in data resources and machine-learning approaches in the MRC and QA systems especially are developed significantly in two resource-rich languages such as English and Chinese. A low-resource language like Vietnamese has witnessed a scarcity of research on QA systems. This paper presents XLMRQA, the first Vietnamese QA system using a supervised transformer-based reader on the Wikipedia-based textual knowledge source (using the UIT-ViQuAD corpus), outperforming the two robust QA systems using deep neural network models: DrQA and BERTserini with 24.46\% and 6.28\%, respectively. From the results obtained on the three systems, we analyze the influence of question types on the performance of the QA systems.

\end{abstract}

\section{Introduction}
In recent years, the rapid development of social media has led to an explosion of data and information. People need to find information and knowledge through the support of machine question answering applications like Google, Siri, and Alexa. QA systems assist people in accessing information and knowledge faster without taking much time and effort. QA-based tasks are of interest to the Vietnamese natural language processing and computational linguistics community. Machine reading comprehension-based QA systems \cite{chen-etal-2017-reading} have gained much attention in recent years. Although several machine reading comprehension corpora are released for developing QA systems such as UIT-ViQuAD \cite{nguyen-etal-2020-vietnamese-dataset}, UIT-ViWikiQA \cite{do2021sentence}, and UIT-ViNewsQA \cite{DBLP:journals/corr/abs-2006-11138}, there is no reader-based QA system for Vietnamese yet.

\begin{figure}
    \centering
    \includegraphics[width=0.6\textwidth]{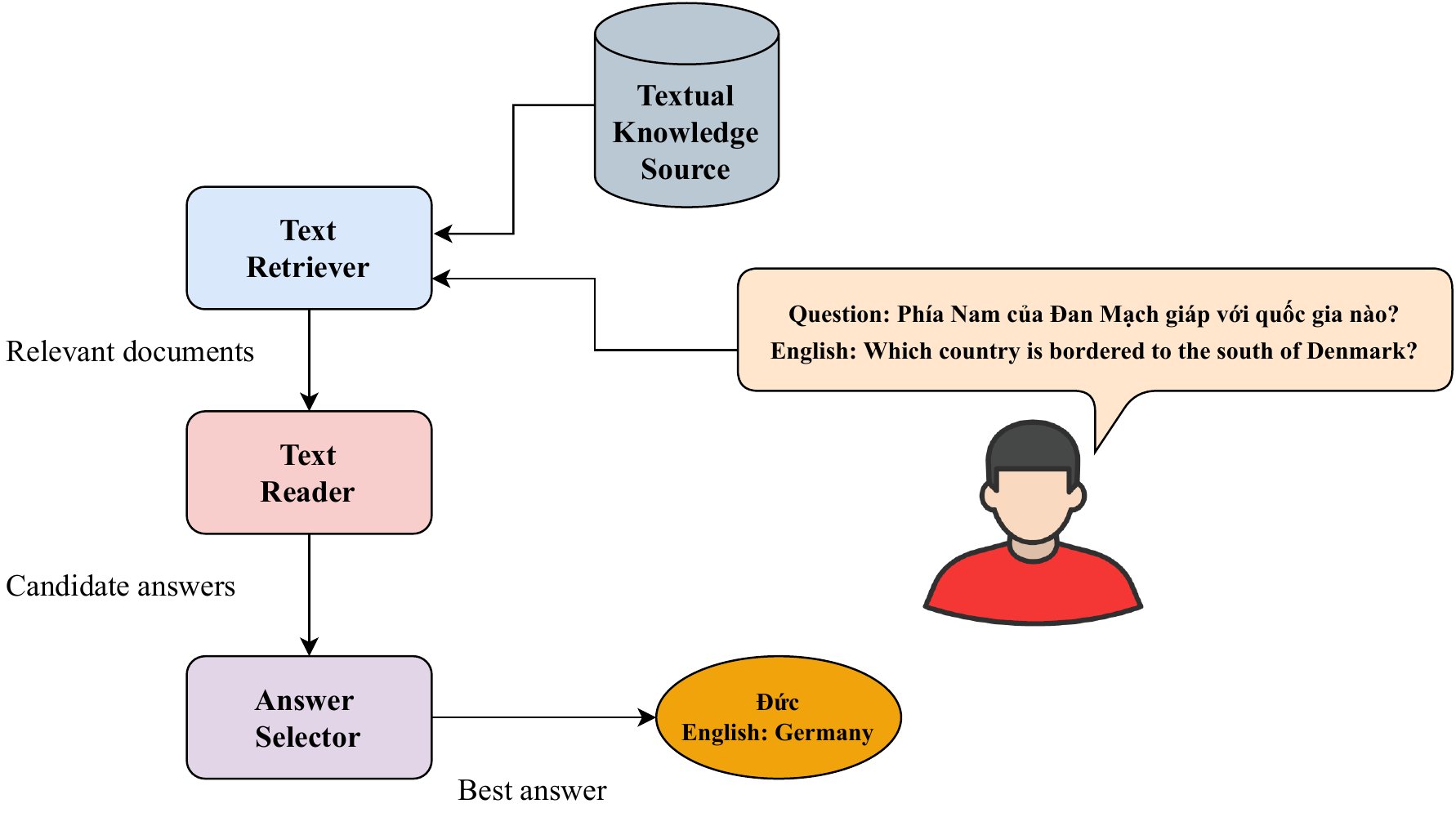}
    \caption{An example of a Retriever-Reader-Selector QA system on the Wikipedia.}
    \label{fig:QA}
\end{figure}

Along with the strong development of machine learning, QA systems have been explored in various corpora and methods. In recent years, QA systems have followed two Retriever-Reader QA systems such as DrQA \cite{chen-etal-2017-reading} and BERTserini \cite{yang-etal-2019-end-end}, respectively. The input of QA systems is a question and a collection of passages or documents, and the output is a predicted answer extracted from a relevant document. Figure \ref{fig:QA} shows the input and output of a QA system on the Vietnamese Wikipedia.

Our three main contributions are described as follows. 
\begin{itemize}
 \item We re-implement state-of-the-art QA systems on the Vietnamese Wikipedia knowledge texts: DrQA \cite{chen-etal-2017-reading} and BERTserini \cite{yang-etal-2019-end-end} as baseline systems. The first experiments were performed on the retriever-reader-based QA model on Vietnamese texts.
\item We propose XLMRQA, a retriever-reader-selector QA system for the Vietnamese language, outperforming the F1-score and exact match (EM) of two other SOTA systems: DrQA with the multi-layer recurrent neural network-based reader and BERTserini with the BERT based reader.
\item We analyze the impacts of question types on the proposed QA system XLMRQA for Vietnamese, which helps researchers improve the performance of the QA systems in future work.
\end{itemize}

\section{Related Work}
Building a QA system requires integrating two main parts: the text retriever and the text reader, to generate the whole QA system. In this section, we briefly review the techniques related to our work.

{\bf Text Retriever}: In this study, we used two popular and effective techniques: TF-IDF and Pyserini. Term Frequency–Inverse Document Frequency (TF-IDF) is the count-based method that reveals the importance of a token or word to a document or text in a corpus \cite{rajaraman2011mining}. Text retrieval models \cite{hiemstra2000probabilistic}, text summarization models \cite{christian2016single}, and question answering models \cite{chen-etal-2017-reading} have all employed TF-IDF. This approach is used as the initial baseline method for text retriever to compare with other techniques. Pyserini is a simple Python package that aids researchers in reproducing their findings by offering excellent first-component document retrieval for multi-component rating systems \cite{DBLP:journals/corr/abs-2102-10073}. Because Pyserini is simple yet effective, it was chosen to be implemented for the text retriever as a component of the QA system.

{\bf Automatic Reader}: The DrQA reader is a multi-layer neural network model that has been trained to identify answers from a document as input. In English \cite{chen-etal-2017-reading}, and Vietnamese \cite{nguyen-etal-2020-vietnamese-dataset,DBLP:journals/corr/abs-2006-11138}, this reader is generated from machine reading comprehension tasks. Text segments from the Pyserini retriever are passed to the BERT reader \cite{yang-etal-2019-end-end}. 
These models were widely applied in automatic reading comprehension tasks in English \cite{chen-etal-2017-reading,devlin-etal-2019-bert} and Vietnamese \cite{nguyen-etal-2020-vietnamese-dataset,DBLP:journals/corr/abs-2006-11138}. Recently, there are more complex models that integrate additional linguistic knowledge into the models \cite{ponti2021minimax,wang2021k}. However, Vietnamese is a language with few resources and does not yet NLP pipeline tools really well to recognize linguistic knowledge to be associated with these language models.

{\bf Full QA System}: Different from the previous QA systems \cite{moldovan2000structure,duong2018hybrid,nguyen2018deep} without readers, DrQA \cite{chen-etal-2017-reading} is a full QA system combining a bigram hash-based TF-IDF retriever with a multi-layer iterative neural network reader trained to predict answers in the passage. BERTserini \cite{yang-etal-2019-end-end} is a QA system that combines the BERT-based reader and the open-source Anserini toolkit for text retriever. The system receives a small set of documents as input. In an end-to-end approach, the system combines best practices from document retrieval with a BERT-based reader to determine answers from a large-scale corpus of English Wikipedia articles. For the Vietnamese language, there are still not any QA systems based on the Retriever-Reader mechanism, mainly focusing on the traditional QA system \cite{bach2020question,le2018factoid}. Therefore, we would like to develop this system as a starting point of the mechanism for Vietnamese QA.

\section{UIT-ViQuAD: Vietnamese Wikipedia-based Textual Knowledge Resource}
In this paper, we use the UIT-ViQuAD corpus (abbreviated as ViQuAD) \cite{nguyen-etal-2020-vietnamese-dataset} as a Wikipedia-based textual knowledge source, which is the main resource to build the text retriever, reader, and the full QA system for Vietnamese.

ViQuAD is a Vietnamese corpus for assessing question answering, machine reading comprehension, and question generation models. Table \ref{tab:overallstatistics} summarizes the statistics on the training (Train), development (Dev), and test (Test) sets of this corpus. ViQuAD comprises over 23,000 triples, and each triple includes a question, its answer, and a passage containing the answer. The numbers of passages and articles, the average question and answer lengths, and lexical unit sizes are presented in Table \ref{tab:overallstatistics}.

\begin{table}[H]
\caption{Overview of the ViQuAD corpus.}
\label{tab:overallstatistics}
\resizebox{\columnwidth}{!}{\begin{tabular}{crrrrrrr}
\hline
\multirow{2}{*}{\textbf{Corpus}} & \multicolumn{1}{c}{\multirow{2}{*}{\textbf{\#article}}} & \multicolumn{1}{c}{\multirow{2}{*}{\textbf{\#passage}}} & \multicolumn{1}{c}{\multirow{2}{*}{\textbf{\#questions}}} & \multicolumn{3}{c}{\textbf{Average length}}                                                                          & \multicolumn{1}{c}{\multirow{2}{*}{\textbf{Vocabulary size}}} \\ \cline{5-7}
                                  & \multicolumn{1}{c}{}                                    & \multicolumn{1}{c}{}                                    & \multicolumn{1}{c}{}                                      & \multicolumn{1}{l}{\textbf{passage}} & \multicolumn{1}{l}{\textbf{question}} & \multicolumn{1}{l}{\textbf{answer}} & \multicolumn{1}{c}{}                                          \\ \hline
Train                             & 138                                                      & 4,101                                                    & 18,579                                                     & \multicolumn{1}{r}{153.9}            & \multicolumn{1}{r}{12.2}              & 8.1                                  & 36,174                                                         \\ 
Dev                               & 18                                                       & 515                                                      & 2,285                                                      & \multicolumn{1}{r}{147.9}            & \multicolumn{1}{r}{11.9}              & 8.4                                  & 9,184                                                          \\ 
Test                              & 18                                                       & 493                                                      & 2,210                                                      & \multicolumn{1}{r}{155.0}            & \multicolumn{1}{r}{12.2}              & 8.9                                  & 9,792                                                          \\ \hline
Full                              & 174                                                      & 5,109                                                    & 23,074                                                     & \multicolumn{1}{r}{153.4}            & \multicolumn{1}{r}{12.2}              & 8.2                                  & 41,773                                                         \\ \hline
\end{tabular}}
\end{table}

Besides, Nguyen et al. \cite{nguyen-etal-2020-vietnamese-dataset} also analyzed the distribution of seven types of questions: Who, What, When, Where, Why, How, and Others in the ViQuAD corpus. The most common type of question is What, accounting for 49.97 percent of all questions, Where questions have the lowest proportion of 5.64 percent, and other types of questions contribute to proportions of between 7 percent and 10 percent.

\begin{figure}
    \centering
    \includegraphics[width=0.6\linewidth]{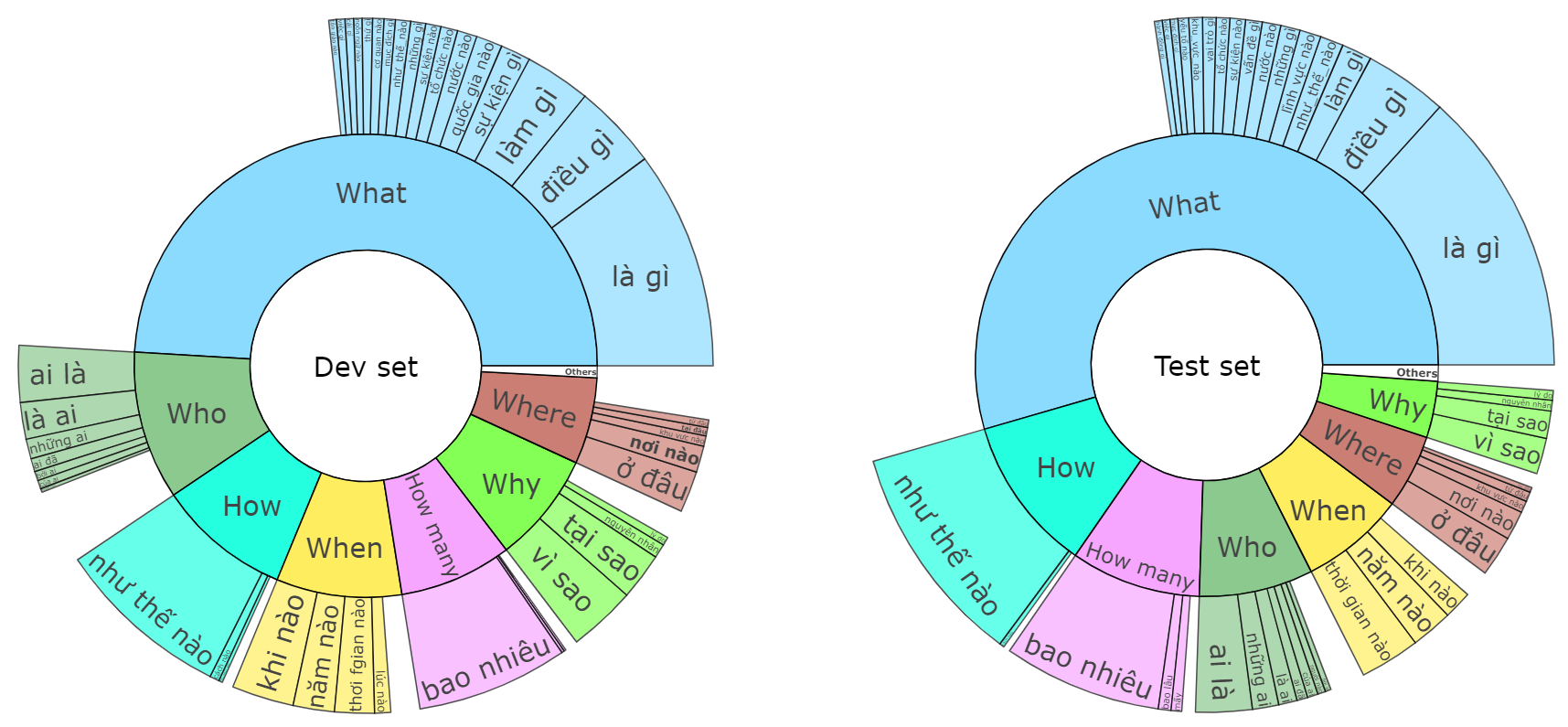}
    \caption{Question types and question words statistics in the corpus \cite{do2021sentence}.}
    \label{fig:words_question}
\end{figure}

The question words in Vietnamese for each question type are diverse. The UIT-ViWiKiQA \cite{do2021sentence} corpus is a reading comprehension corpus that is automatically converted from the ViQuAD corpus's question-answer pairs. Do et al. \cite{do2021sentence} analyzed diverse Vietnamese question words to pose in each question type. Figure \ref{fig:words_question} shows the proportions of types of questions on the Dev and Test sets.

The linguistic phenomenon in Vietnamese questions has various question words in different question types: What, Who, When, Why, and Where. What questions have the most variety of question words compared to other question types. Only the question words with a high frequency of occurrence are presented in Figure \ref{fig:words_question} (approximately 0.7 percent or more). For example, "là gì", "điều gì", "làm gì", and "cái gì" are question words in "What questions" with "là gì" having the most significant rate (24.42 percent in the Test set and 20.69 percent in the Dev set). The question words in the How and How many question types are usually not diversified. In the Dev set, "như thế nào" is the most popular question word reaching at 87.79 percent, whereas it accounts for 95.02 percent in the Test set. According to the previous investigation results \cite{nguyen-etal-2020-vietnamese-dataset}, the What question type contains the most significant proportion of question words and the most remarkable diversity of question terms. The remaining types of questions make up a small percentage of the total, particularly the How and How many question categories, which include few question words.


\section{XMLRQA: Retriever-Reader-Selector Question Answering System for the Vietnamese language}

\begin{figure}[H]
    \centering
    \includegraphics[width=0.6\textwidth]{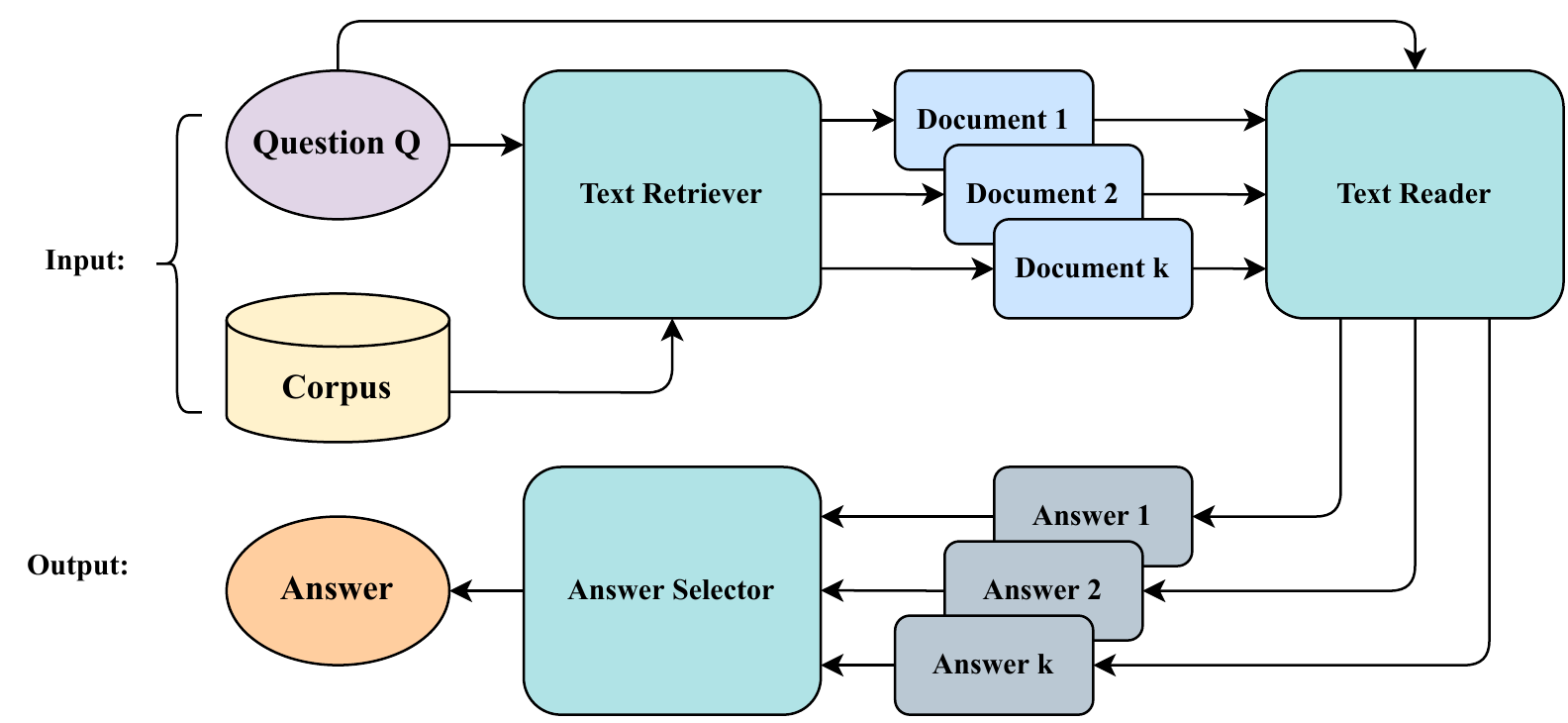}
    \caption{Overview of the Retriever-Reader-Selector QA system for Vietnamese.}
    \label{fig:QA_system}
\end{figure}

\subsection{Overview of QA System Architecture}

Inspired by the DrQA system \cite{chen-etal-2017-reading}, we present an overview of the QA architecture using a supervised reader for the Vietnamese language. Figure \ref{fig:QA_system} presents a QA system with three components: text retriever, text reader, and answer selector. The text retriever finds texts or documents and passes them to the text reader to find the candidate answers. The answer selector finds the answer that best matches the question from the candidate answers predicted by the text reader. In particular, we describe the QA system and its components as follows.

\subsection{Text Retriever}
\label{Document_retriever}
We apply a basic retriever to find k passages that answer the input question, using the question as a bag-of-words question. The text retriever finds passages or documents related to the question from a set of 5,109 passages extracted from the ViQuAD corpus \cite{nguyen-etal-2020-vietnamese-dataset}. This corpus was built by aggregating all the passages from the Train, Dev, and Test sets of the ViQuAD corpus consisting of 4,101, 515, and 493 passages, respectively. This paper assesses two different text retrievers, including TF-IDF and the Anserini. To optimize the performance of QA systems, we apply word segmentation to the text retrievers.

\subsection{Text Reader}
\label{Document_Reader}
The retrieved passages are passed to the reader to extract the candidate answers. The questions are combined with their passages to generate k question-passage pairs that enter the reader to predict k candidate answers.

For each question-passage pair that enters the reader, the model reads the question with M tokens [$q_1$, $q_2$, ..., $q_M$] and then reads all N tokens in the passage [$d_1$, $d_2$, $d_N$]. The model performs two probabilities for each token $d_i$ in the document, with $Pstart_i$ being the score when $d_i$ is the starting answer position and $Pend_i$ being the score when $d_i$ is the ending answer position. The reader selects the best answer span from the $Pstart$ and $Pend$ scores list with the highest score from the document. After processing k passages by the reader, the reader obtains k answers and the answer-score list.


\begin{figure}[H]
    \centering
    \includegraphics[width=0.5\textwidth]{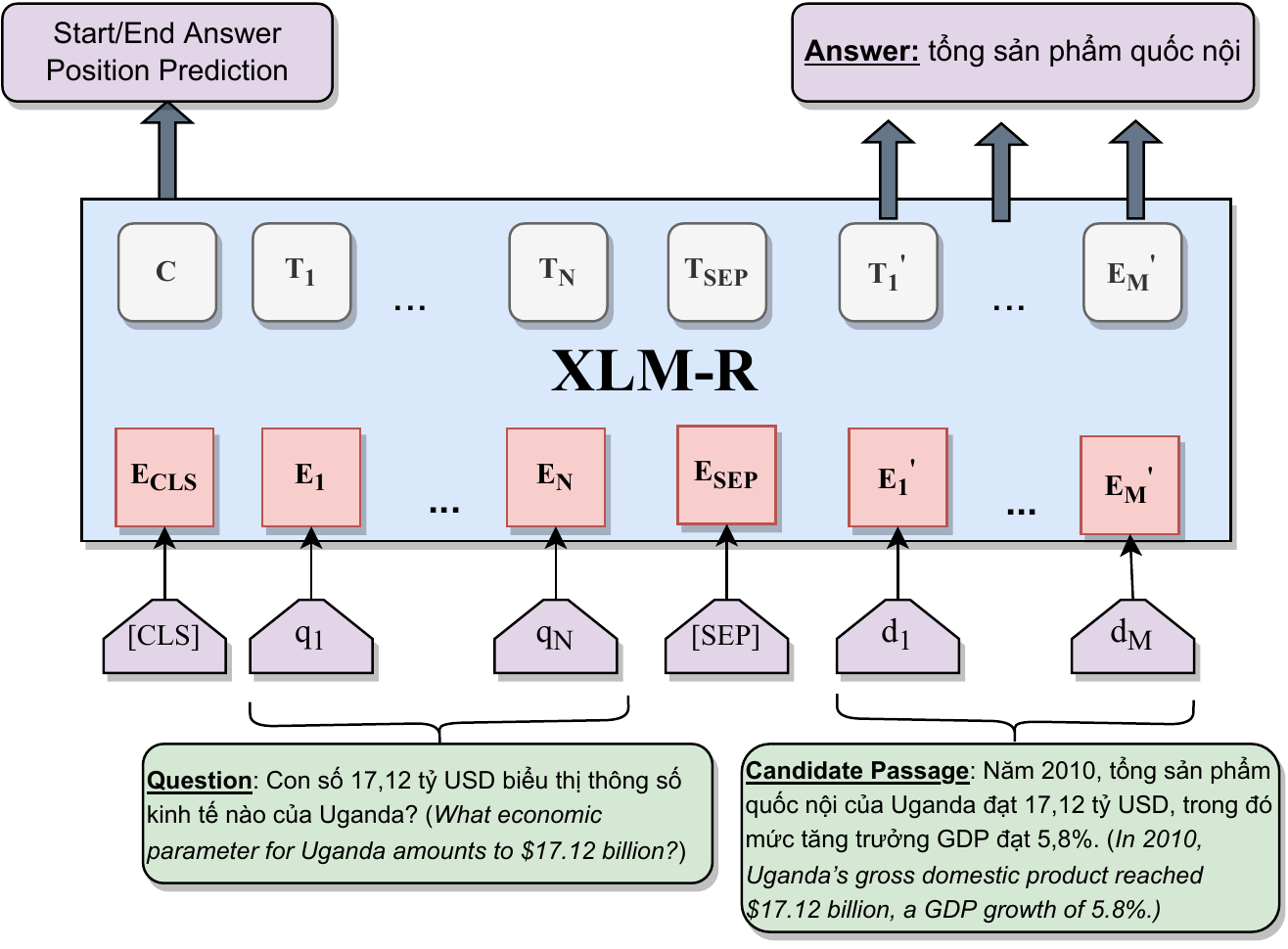}
    \caption{Overview of the XLM-R based Reader for the Vietnamese language.}
    \label{fig:QA_system}
\end{figure}

XLM-RoBERTa (XLM-R) \cite{conneau-etal-2020-unsupervised} is a multilingual language model trained on a large-scale dataset with 100 languages. XLM-R is used as a pre-trained language model for many tasks such as natural language inference and machine reading comprehension, which achieves state-of-the-art performances. In this paper, we use XLM-R to build a reader as the main component of the XLMRQA system to extract candidate answers before transferring them into the answer selector. Figure \ref{fig:QA_system} shows an overview of the XLM-R-based Reader for the Vietnamese language.

\subsection{Answer Selector}
\label{answer_selection}
The candidate answer list, the reading score list, and the retrieving score list are fed into this component. Each score in the two score lists corresponds to each answer in the answer list. Following Yang et al. \cite{yang-etal-2019-end-end}, we then combine the reading score with the retrieving score through linear interpolation to estimate the score for each answer and find the answer with the highest score.

\begin{equation}
    \label{CT1}
    Score_{answer} = alpha*Score_{Reader}+(1-alpha)*Score_{Retriever}
\end{equation}

where alpha is a hyper-parameter whose value is in the range [0,1], and alpha is found by tuning with a thousand question-answer pairs extracted randomly from the Train set.

\section{Experimental Evaluation}
\subsection{Baseline Systems}

For baseline systems, we re-implement two QA systems that have achieved state-of-the-art performance on English and Chinese corpora, including DrQA \cite{chen-etal-2017-reading} and BERTserini \cite{yang-etal-2019-end-end}. Also, we compare the two QA systems with our proposed system using a powerful XLM-R-based reader, which obtains the best performance on the machine reading comprehension task on the ViQuAD corpus \cite{nguyen-etal-2020-vietnamese-dataset}.

\subsection{Experimental Settings} We used a single
NVIDIA Tesla P100 GPU on the Google Collaboratory server\footnote{https://colab.research.google.com/} for all our experiments. We use sqlite3\footnote{https://docs.python.org/3/library/sqlite3.html} to store the aggregated corpus from the ViQuAD corpus \cite{nguyen-etal-2020-vietnamese-dataset}. We set up our experiments described as follows.

\textbf{Text retriever:} We implement two models for text retrievers, including TF-IDF and Anserini. The TF-IDF model is based on the DrQA model with the bi-gram language model. The Anserini model is a Python-compatible version called Pyserini\footnote{https://github.com/castorini/pyserini}. In addition, in analyzing the influence of word splitting on text retrievers, we use pyvi\footnote{https://pypi.org/project/pyvi/} as a word segmentation tool.

\textbf{Text reader:} We implement DrQA reader based on the DrQA system \cite{chen-etal-2017-reading} as text reader as the first model and trained through 30 epochs with batch-size = 32. The pre-trained word embedding model when implementing the DrQA model is ELMO \cite{peters-etal-2018-deep,vu2019etnlp}. We use powerful pre-trained language models: BERT \cite{devlin-etal-2019-bert} and XLM-R \cite{conneau-etal-2020-unsupervised} as text readers, of which the XLM-R model with two versions: large and base. The BERT and XLM-R models are fine-tuned with the baseline configurations provided by Huggingface\footnote{https://huggingface.co/} and we set them with a number of epochs = 2, a maximum string length = 384, and the learning rate = 2e-5.

\subsection{Experimental Results}
We assess the performance of each component, the end-to-end QA system and analyze the experimental results.

\subsubsection{Experimental Results of Text Retriever\newline}
First and foremost, all the passages in the corpus are indexed. The text retriever then provides a score value to each passage, representing the likelihood that the passage includes the answer. Text retriever selects k passages with the highest score. P@k is used to assess the text retriever and is defined as the ratio of ranked passages that contained the answers to the top K relevant passages. Given $Q=\{q_1, q_2,..., q_n\}$ as a set of questions, the answer of $q \in Q$ is $a_q$, $P \in Ps$ is the passage containing $a_q$ where $Ps$ if the set of passages in the corpus, and $P_k^*(q) \subset Ps$ is the set of the top k passages predicted by the text retriever. P@k is calculated using Formula \ref{P@k}.

\begin{equation}
    \label{P@k}
    P@k=\frac{1}{|Q|}\sum_{1}^{n}
    \begin{cases}
    1 & \text{if $P \in P_k^*(q)$},\\
    0 & \text{Otherwise.}
    \end{cases}
\end{equation}

Table \ref{tab:resultDR} shows the P@k of two text retrievers, and it shows the effect of word segmentation for two text retrievers. Adding a word segmentation makes the TF-IDF model less efficient as its P@k decreases overall instances of k, decreasing by 5.60\% on average. Pyserini model becomes more efficient when combined with pyvi as a word segmentation tool. Overall, Pyserini increases accuracy on all instances of k, increasing by 1.04\% on average. Thus, when building the QA systems, we use the word segmentation for Pyserini and not for TF-IDF.

\begin{table}[H]
\centering
\caption{P@k of the text retriever on the Test set of the ViQuAD corpus.}
\label{tab:resultDR}
\begin{tabular}{lcccc}
\hline
     & \textbf{TF-IDF} & \textbf{TF-IDF+Pyvi}   & \textbf{Pyserini} & \textbf{Pyserini+Pyvi} \\ \hline
P@1  & 64.39  & 58.14 (-6.25) & 64.89    & 68.51 (+3.62) \\
P@5  & 84.34  & 78.42 (-5.92) & 85.52    & 86.47 (+0.95) \\
P@10 & 90.68  & 84.89 (-5.79) & 89.95    & 91.22 (+1.27) \\
P@15 & 92.99  & 87.96 (-5.03) & 92.08    & 93.08 (+1.00) \\
P@20 & 94.39  & 88.82 (-5.57) & 93.76    & 93.98 (+0.22) \\
P@25 & 95.38  & 90.09 (-5.29) & 94.71    & 94.76 (+0.05) \\
P@30 & 96.06  & 90.72 (-5.34) & 95.25    & 95.45 (+0.20) \\ \hline
\end{tabular}
\end{table}

\subsubsection{Experimental Results of Text Reader\newline}

Text reader is used for extracting candidate answers from questions and their passages obtained by text retrievers. The text reader receives a question and a passage as input of the reader. For the reading component, exact match (EM) and F1 (following the evaluation of the ViQuAD corpus \cite{nguyen-etal-2020-vietnamese-dataset}) are the two assessment measures employed to estimate the performance of the text reader.

\begin{table}[]
\centering
\caption{Evaluation of text readers on the ViQuAD corpus.}
\label{tab:resultRD}
\begin{tabular}{lcccc}
\hline
                     & \multicolumn{2}{c}{\textbf{Dev}}   & \multicolumn{2}{c}{\textbf{Test}}  \\ \cline{2-5} 
                     & \textbf{EM}    & \textbf{F1} & \textbf{EM}    & \textbf{F1} \\ \hline
DrQA reader          & 44.60          & 65.99             & 39.10          & 62.92             \\
mBERT                & 63.30          & 80.69             & 61.59          & 80.87             \\
XLM-R$_{Base}$           & 63.60          & 81.95             & 63.87          & 82.56             \\
\textbf{XLM-R$_{Large}$} & \textbf{73.23} & \textbf{88.36}    & \textbf{70.29} & \textbf{86.86}    \\ \hline
\end{tabular}
\end{table}

The performance of the text readers on the Dev and Test sets of the ViQuAD corpus is shown in Table \ref{tab:resultRD}. The BERT and XLM-R models outperform the DrQA Reader model in terms of overall performance. On the Test set of the ViQuAD corpus, the XLMR$_{Large}$ model outperforms the other models in both evaluation metrics, with an F1 of 86.86 percent and an EM of 70.29 percent.

\subsubsection{Experimental Results of Full QA Systems\newline}

The QA system is a complete system with three modules: text retriever, text reader, and answer selector. We implement three QA systems, including DrQA \cite{chen-etal-2017-reading}, BERTserini \cite{yang-etal-2019-end-end}, and XLMRQA, where DrQA's text retriever is TF-IDF, and the other two are Pyserini. We use the DrQA reader for the DrQA system, BERT \cite{devlin-etal-2019-bert} for the reader of BERTserini, and XLM-R$_{Large}$ \cite{conneau-etal-2020-unsupervised} for the reader of XLMRQA. We use the two assessment metrics including EM and F1 scores to measure QA systems' performance.

The evaluation of the QA systems is shown in Table \ref{tab:resultThreeQASystems}. The three systems in ascending order of results are DrQA, BERTserini, and XLMRQA. The XLMRQA QA system achieves the highest performance with EM and F1 set to the maximum at k=5 with an F1 of 64.99\% and an EM of 51.94\% on the Test set. Similar to XLMRQA, the BERTserini system achieves the best performance at k=5 with an F1 of 58.30\% and an EM of 39.46\% on the ViQUAD Test set. The DrQA system achieves the best performance with $k\geq10$, and these are equal. Nevertheless, we have chosen k=10 as the official value for the DrQA system because it achieves good performance in terms of time. At k=10, the DrQA system achieved an F1 of 37.86\% and EM of 18.42\% on the Test set.

\begin{table}[]
\centering
\caption{Performances of the Vietnamese QA systems with different k-values.}
\label{tab:resultThreeQASystems}
\begin{tabular}{lcccccccccccc}
\hline
\multirow{3}{*}{\textbf{k}} & \multicolumn{4}{c}{\textbf{DrQA}}                                       & \multicolumn{4}{c}{\textbf{BERTserini}}                                 & \multicolumn{4}{c}{\textbf{XLMRQA}}                                 \\ \cline{2-13} 
                            & \multicolumn{2}{c}{\textbf{Dev}}   & \multicolumn{2}{c}{\textbf{Test}}  & \multicolumn{2}{c}{\textbf{Dev}}   & \multicolumn{2}{c}{\textbf{Test}}  & \multicolumn{2}{c}{\textbf{Dev}}   & \multicolumn{2}{c}{\textbf{Test}}  \\ \cline{2-13} 
                            & \textbf{EM}    & \textbf{F1} & \textbf{EM}    & \textbf{F1} & \textbf{EM}    & \textbf{F1} & \textbf{EM}    & \textbf{F1} & \textbf{EM}    & \textbf{F1} & \textbf{EM}    & \textbf{F1} \\ \hline
\textbf{1}                  & 18.42          & 37.17             & 17.87          & 37.37             & 38.07          & 53.89             & 36.52          & 55.55             & 47.96          & 60.39             & 47.96          & 61.83             \\
\textbf{5}                  & 19.17          & 38.05             & 18.37          & 37.86             & \textbf{41.84} & \textbf{57.50}    & \textbf{39.46} & \textbf{58.30}    & \textbf{51.99} & \textbf{64.10}    & \textbf{51.94} & \textbf{64.99}    \\
\textbf{10}                 & \textbf{19.17} & \textbf{38.05}    & \textbf{18.42} & \textbf{37.86}    & 41.75          & 57.21             & 39.41          & 57.98             & 51.77          & 63.79             & 51.36          & 64.49              \\
\textbf{15}                 & 19.17          & 38.04             & 18.42          & 37.86             & 41.71          & 57.10             & 39.50          & 58.09             & 51.77          & 63.79             & 51.36          & 64.49             \\
\textbf{20}                 & 19.17          & 38.05             & 18.42          & 37.86             & 41.71          & 57.09             & 39.46          & 58.03             & 51.77          & 63.79             & 51.36          & 64.49             \\
\textbf{25}                 & 19.17          & 38.05             & 18.42          & 37.86             & 41.71          & 57.08             & 39.50          & 58.00             & 51.77          & 63.79             & 51.36          & 64.49             \\
\textbf{30}                 & 19.17          & 38.05             & 18.42          & 37.86             & 41.71          & 57.08             & 39.50          & 57.99             & 51.77          & 63.79             & 51.36          & 64.49             \\ \hline
\end{tabular}
\end{table}

\subsection{Result Analysis}

\begin{figure}
    \centering
    \includegraphics[width=0.6\textwidth]{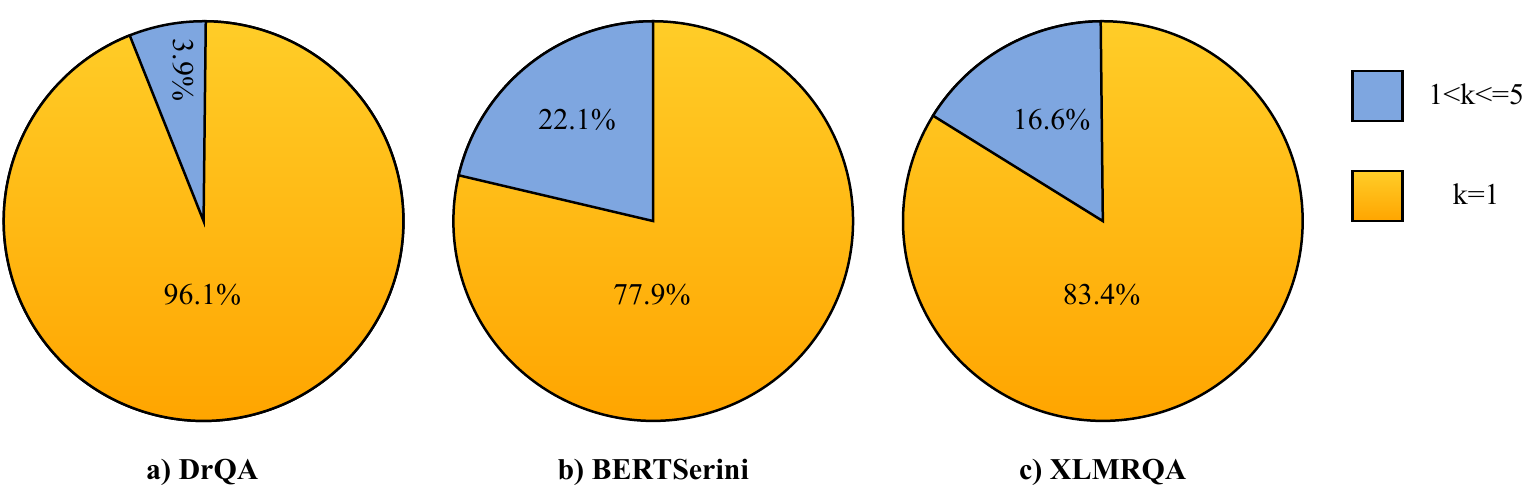}
    \caption{Distribution of where predicted answers appear in the set of k passages.}
    \label{fig:PositionofAnswer}
\end{figure}

The majority of the predicted answers are focused on the first five passages, as seen in Figure \ref{fig:PositionofAnswer}. For example, the XLMRQA system shows that 83.40\% of the questions obtain predicted answers correctly when retrieving the passage with k=1, and all questions with correct answers appear in the top five passages. This explains the results in Table \ref{tab:resultThreeQASystems} that when $k\geq10$, the results on the Test set are almost unchanged on both F1 and EM. The same is true for DrQA and BERTserini systems. Up to 99.90\% of the answers appear the first to fifth passages in the DrQA system. This makes the results on the Test set of the DrQA system unchanged with $k\geq10$.

\begin{figure}
    \centering
    \includegraphics[width=0.85\textwidth]{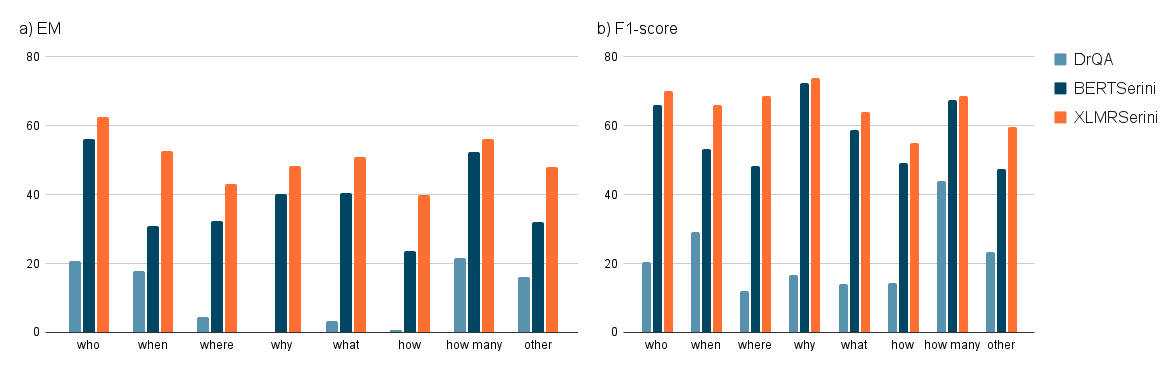}
    \caption{Evaluation of QA systems on question types of the test set of ViQuAD.}
    \label{fig:ResultQuestionType}
\end{figure}

Figure \ref{fig:ResultQuestionType} shows the performance of QA systems for different types of questions. Overall, the XLMRQA QA system achieved the highest performance based on EM and F1 for all question types. The second-highest performing system is BERTserini, and the last is DrQA. Our analysis shows that the amount and diversity of question words and the complexity of the question impacted the performance of QA systems for each question type. Because Why and How are challenging to answer because they demand that the system comprehends the question and the retrieved passage, QA systems do not perform well. When questions include a limited number of questions but a large variety of question words (see Figure \ref{fig:words_question}) makes it difficult for QA systems, resulting in poor performance. Although the What type of question accounts for a vast number (nearly 50\%) but has the highest diversity of words to ask, the QA system is problematic in recognizing and extracting answers. As for the How many question type, there is a shallow diversity of words to ask (see Figure \ref{fig:words_question}), so the performance of QA systems is higher. Analyzing the performance of QA systems based on question types to assess the difficulty level of Vietnamese questions for the QA task in ViQuAD helps researchers explore better models for Vietnamese in the future.

\begin{figure}
    \centering
    \includegraphics[width=0.4\textwidth]{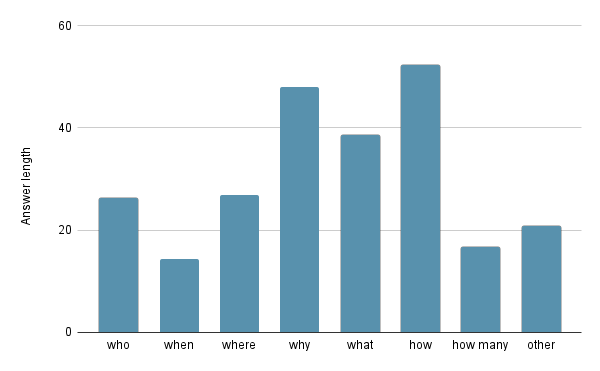}
    \caption{Average predicted answer length for each question type.}
    \label{fig:answerlength}
\end{figure}

Figure \ref{fig:answerlength} shows the average answer length for each question type. We found that the answers to the three types of What, Why, and How questions have the largest length. Thus, the performances of the QA systems on these three question types achieve low EM scores but high F1 scores. Especially with the Why question type, although QA systems achieve deficient performance on EM, QA systems achieve high performance on F1 because the answers to this type of question are long. This shows that the evaluation of the QA task based on F1 is only relative because it is easily dependent on the answer length.

\textbf{How many languages will our proposed system implement?} A powerful pre-trained language model XLM-RoBERTa using transformer architecture supports 100 different languages (including low-resource languages). Based on our guidelines for building QA systems with the Retriever-Reader-Selector mechanism in this paper, QA systems for other languages (especially low-resource languages) can be easily adapted and re-implemented as baseline QA systems. This proposed system can be extended to different datasets on other languages such as KorQuAD (for Korean) \cite{lim2019korquad1}, SberQuAD (for Russian) \cite{efimov2020sberquad}, JaQuAD (for Japanese) \cite{so2022jaquad}, and FQuAD (for French) \cite{d2020fquad} in the near future.

\section{Conclusion and Future Work}

In this paper, we introduced XLMRQA, a QA system based on the retriever-reader-selector mechanism for Vietnamese open-domain texts, which outperformed two state-of-the-art question answering systems, DrQA \cite{chen-etal-2017-reading}, and BERTserini \cite{yang-etal-2019-end-end}, on the ViQuAD corpus \cite{nguyen-etal-2020-vietnamese-dataset}. For assessing the performance of three QA systems on the ViQuAD corpus, we achieved the highest performance with the XLMRQA system: EM of 51.94\% and F1 of 64.99\%. Analysis of the performance of QA systems was performed on different types of questions. The results of our analysis indicate that the types of challenging questions to be addressed in future studies are How, Why, Where, and What. In the future, several future directions are recommended: (1) integrating diverse question words as linguistic features into the QA systems can boost their performances; (2) finding out a method to leverage the power of monolingual and multilingual BERTology-based language models \cite{rogers2020primer}; and (3) expanding our QA system to other low-resource languages.

\section*{Acknowledgement}

This research is funded by University of Information Technology-Vietnam National University HoChiMinh City under grant number D1-2022-13. Kiet Van Nguyen was funded by Vingroup JSC and supported by the Master, PhD Scholarship Programme of Vingroup Innovation Foundation (VINIF), Institute of Big Data, code VINIF.2021.TS.026.

\bibliographystyle{plain} 
\bibliography{references} 

\begin{thebibliography}{10}

\bibitem{bach2020question}
Ngo~Xuan Bach, Phan~Duc Thanh, and Tran~Thi Oanh.
\newblock Question analysis towards a vietnamese question answering system in
  the education domain.
\newblock {\em Cybernetics and Information Technologies}, 20(1):112--128, 2020.

\bibitem{chen-etal-2017-reading}
Danqi Chen, Adam Fisch, Jason Weston, and Antoine Bordes.
\newblock Reading {W}ikipedia to answer open-domain questions.
\newblock In {\em Proceedings of ACL 2017 (Volume 1: Long Papers)}, pages
  1870--1879, Vancouver, Canada, July 2017. Association for Computational
  Linguistics.

\bibitem{christian2016single}
Hans Christian, Mikhael~Pramodana Agus, and Derwin Suhartono.
\newblock Single document automatic text summarization using term
  frequency-inverse document frequency (tf-idf).
\newblock {\em ComTech: Computer, Mathematics and Engineering Applications},
  7(4):285--294, 2016.

\bibitem{conneau-etal-2020-unsupervised}
Alexis Conneau, Kartikay Khandelwal, Naman Goyal, Vishrav Chaudhary, Guillaume
  Wenzek, Francisco Guzm{\'a}n, Edouard Grave, Myle Ott, Luke Zettlemoyer, and
  Veselin Stoyanov.
\newblock Unsupervised cross-lingual representation learning at scale.
\newblock In {\em Proceedings of ACL 2020}, pages 8440--8451, Online, July
  2020. Association for Computational Linguistics.

\bibitem{devlin-etal-2019-bert}
Jacob Devlin, Ming-Wei Chang, Kenton Lee, and Kristina Toutanova.
\newblock Bert: Pre-training of deep bidirectional transformers for language
  understanding.
\newblock In {\em Proceedings of NAACL 2019, Volume 1 (Long and Short Papers)},
  pages 4171--4186, 2019.

\bibitem{do2021sentence}
Phong Nguyen-Thuan Do, Nhat~Duy Nguyen, Tin~Van Huynh, Kiet~Van Nguyen, Anh
  Gia-Tuan Nguyen, and Ngan Luu-Thuy Nguyen.
\newblock Sentence extraction-based machine reading comprehension for
  vietnamese.
\newblock In {\em International Conference on Knowledge Science, Engineering
  and Management}, pages 511--523. Springer, 2021.

\bibitem{duong2018hybrid}
Phuc~H Duong, Hien~T Nguyen, Duy~D Nguyen, and Hao~T Do.
\newblock A hybrid approach to answer selection in question answering systems.
\newblock In {\em International Symposium on Integrated Uncertainty in
  Knowledge Modelling and Decision Making}, pages 191--202. Springer, 2018.

\bibitem{d2020fquad}
Martin d’Hoffschmidt, Wacim Belblidia, Quentin Heinrich, Tom Brendl{\'e}, and
  Maxime Vidal.
\newblock Fquad: French question answering dataset.
\newblock In {\em Findings of the Association for Computational Linguistics:
  EMNLP 2020}, pages 1193--1208, 2020.

\bibitem{efimov2020sberquad}
Pavel Efimov, Andrey Chertok, Leonid Boytsov, and Pavel Braslavski.
\newblock Sberquad--russian reading comprehension dataset: Description and
  analysis.
\newblock In {\em International Conference of the Cross-Language Evaluation
  Forum for European Languages}, pages 3--15. Springer, 2020.

\bibitem{hiemstra2000probabilistic}
Djoerd Hiemstra.
\newblock A probabilistic justification for using tf$\times$ idf term weighting
  in information retrieval.
\newblock {\em International Journal on Digital Libraries}, 3(2):131--139,
  2000.

\bibitem{le2018factoid}
Phuong Le-Hong and Duc-Thien Bui.
\newblock A factoid question answering system for vietnamese.
\newblock In {\em Companion Proceedings of the The Web Conference 2018}, pages
  1049--1055, 2018.

\bibitem{lim2019korquad1}
Seungyoung Lim, Myungji Kim, and Jooyoul Lee.
\newblock Korquad1. 0: Korean qa dataset for machine reading comprehension.
\newblock {\em arXiv preprint arXiv:1909.07005}, 2019.

\bibitem{DBLP:journals/corr/abs-2102-10073}
Jimmy Lin, Xueguang Ma, Sheng-Chieh Lin, Jheng-Hong Yang, Ronak Pradeep, and
  Rodrigo Nogueira.
\newblock Pyserini: An easy-to-use python toolkit to support replicable ir
  research with sparse and dense representations.
\newblock {\em arXiv preprint arXiv:2102.10073}, 2021.

\bibitem{moldovan2000structure}
Dan Moldovan, Sanda Harabagiu, Marius Pasca, Rada Mihalcea, Roxana Girju,
  Richard Goodrum, and Vasile Rus.
\newblock The structure and performance of an open-domain question answering
  system.
\newblock In {\em Proceedings of ACL 2000}, pages 563--570, 2000.

\bibitem{nguyen-etal-2020-vietnamese-dataset}
Kiet~Van Nguyen, Duc-Vu Nguyen, Anh Gia-Tuan Nguyen, and Ngan Luu-Thuy Nguyen.
\newblock A {V}ietnamese dataset for evaluating machine reading comprehension.
\newblock In {\em Proceedings of the 28th International Conference on
  Computational Linguistics}, pages 2595--2605, Barcelona, Spain (Online),
  December 2020. International Committee on Computational Linguistics.

\bibitem{nguyen2018deep}
Van-Tu Nguyen and Anh-Cuong Le.
\newblock Deep neural network-based models for ranking question-answering pairs
  in community question answering systems.
\newblock In {\em International Symposium on Integrated Uncertainty in
  Knowledge Modelling and Decision Making}, pages 179--190. Springer, 2018.

\bibitem{peters-etal-2018-deep}
Matthew~E. Peters, Mark Neumann, Mohit Iyyer, Matt Gardner, Christopher Clark,
  Kenton Lee, and Luke Zettlemoyer.
\newblock Deep contextualized word representations.
\newblock In {\em Proceedings of NAACL 2018, Volume 1 (Long Papers)}, pages
  2227--2237, New Orleans, Louisiana, June 2018. ACL.

\bibitem{ponti2021minimax}
Edoardo~Maria Ponti, Rahul Aralikatte, Disha Shrivastava, Siva Reddy, and
  Anders S{\o}gaard.
\newblock Minimax and neyman--pearson meta-learning for outlier languages.
\newblock In {\em Findings of ACL: ACL-IJCNLP 2021}, pages 1245--1260, 2021.

\bibitem{rajaraman2011mining}
Anand Rajaraman and Jeffrey~David Ullman.
\newblock {\em Mining of massive datasets}.
\newblock Cambridge University Press, 2011.

\bibitem{rogers2020primer}
Anna Rogers, Olga Kovaleva, and Anna Rumshisky.
\newblock A primer in bertology: What we know about how bert works.
\newblock {\em TACL}, 8:842--866, 2020.

\bibitem{so2022jaquad}
ByungHoon So, Kyuhong Byun, Kyungwon Kang, and Seongjin Cho.
\newblock Jaquad: Japanese question answering dataset for machine reading
  comprehension.
\newblock {\em arXiv preprint arXiv:2202.01764}, 2022.

\bibitem{DBLP:journals/corr/abs-2006-11138}
Kiet Van~Nguyen, Tin Van~Huynh, Duc-Vu Nguyen, Anh Gia-Tuan Nguyen, and Ngan
  Luu-Thuy Nguyen.
\newblock New vietnamese corpus for machine reading comprehension of health
  news articles.
\newblock {\em arXiv preprint arXiv:2006.11138}, 2020.

\bibitem{vu2019etnlp}
Xuan-Son Vu, Thanh Vu, Son~N Tran, and Lili Jiang.
\newblock Etnlp: A visual-aided systematic approach to select pre-trained
  embeddings for a downstream task.
\newblock {\em arXiv preprint arXiv:1903.04433}, 2019.

\bibitem{wang2021k}
Ruize Wang, Duyu Tang, Nan Duan, Zhongyu Wei, Xuan-Jing Huang, Jianshu Ji,
  Guihong Cao, Daxin Jiang, and Ming Zhou.
\newblock K-adapter: Infusing knowledge into pre-trained models with adapters.
\newblock In {\em Findings of ACL: ACL-IJCNLP 2021}, pages 1405--1418, 2021.

\bibitem{yang-etal-2019-end-end}
Wei Yang, Yuqing Xie, Aileen Lin, Xingyu Li, Luchen Tan, Kun Xiong, Ming Li,
  and Jimmy Lin.
\newblock End-to-end open-domain question answering with {BERT}serini.
\newblock In {\em Proceedings of NAACL 2019 (Demonstrations)}, pages 72--77,
  Minneapolis, Minnesota, June 2019. Association for Computational Linguistics.

\end{thebibliography}

\end{document}